\documentclass[sigconf]{acmart}

\usepackage{amsmath}
\usepackage{graphicx}
\usepackage{listings}
\usepackage{xcolor}
\definecolor{codegreen}{rgb}{0,0.6,0}
\definecolor{codegray}{rgb}{255,0,0}
\definecolor{codered}{rgb}{255,0,0}
\definecolor{codepurple}{rgb}{0.58,0,0.82}
\definecolor{backcolour}{rgb}{1,1,1}
\definecolor{mygreen}{HTML}{3E9252}

\lstdefinestyle{mystyle}{
    backgroundcolor=\color{backcolour},   
    commentstyle=\color{white!40!black},
    keywordstyle=\color{magenta},
    numberstyle=\ttfamily\footnotesize\color{codegray},
    stringstyle=\color{codepurple},
    basicstyle=\ttfamily\footnotesize,
    breakatwhitespace=false,         
    breaklines=true,                 
    captionpos=b,                    
    keepspaces=true,                 
    numbers=left,                    
    showspaces=false,                
    showstringspaces=false,
    morekeywords={torch,self},
    keywordstyle=\bfseries\color{mygreen},
    showtabs=false,                  
    tabsize=2,
    xleftmargin=1.3\parindent,
    aboveskip=5pt, 
    belowcaptionskip=5pt,
}

\newcommand{\R}{\mathbb{R}}

\copyrightyear{2023}
\acmYear{2023}
\setcopyright{licensedusgovmixed}\acmConference[SC-W 2023]{Workshops of
The International Conference on High Performance Computing, Network,
Storage, and Analysis}{November 12--17, 2023}{Denver, CO, USA}
\acmBooktitle{Workshops of The International Conference on High Performance
Computing, Network, Storage, and Analysis (SC-W 2023), November 12--17,
2023, Denver, CO, USA}
\acmPrice{15.00}
\acmDOI{10.1145/3624062.3624257}
\acmISBN{979-8-4007-0785-8/23/11}

\begin{document}

\title{Benchmarking and In-depth Performance Study of Large Language Models on Habana Gaudi Processors}

\settopmatter{authorsperrow=4}

\newcommand{\iu}{Indiana University}
\newcommand{\sit}{Stevens Institute of Technology}
\newcommand{\anl}{Argonne National Lab}

\newcommand{\AFFIL}[4]{%
    \affiliation{
        \institution{\small #1}
        \city{#2}\state{#3}\country{#4}
    }
    }

\newcommand{\IU}{\AFFIL{\iu}{Bloomington}{IN}{USA}}
\newcommand{\SIT}{\AFFIL{\sit}{Hoboken}{NJ}{USA}}
\newcommand{\ANL}{\AFFIL{\anl}{Lemont}{IL}{USA}}

\author{Chengming Zhang}{\IU}
\email{czh5@iu.edu}

\author{Baixi Sun}{\IU}
\email{sunbaix@iu.edu}

\author{Xiaodong Yu}{\SIT}
\email{xyu38@stevens.edu}

\author{Zhen Xie}{\ANL}
\email{zhen.xie@anl.gov}

\author{Weijian Zheng}{\ANL}
\email{wzheng@anl.gov}

\author{Kamil Iskra}{\ANL}
\email{iskra@anl.gov}

\author{Pete Beckman}{\ANL}
\email{beckman@anl.gov}

\author{Dingwen Tao}{\IU}
\email{ditao@iu.edu}

\renewcommand{\shortauthors}{Zhang et al.}

\begin{abstract}
Transformer models have achieved remarkable success in various machine learning tasks but suffer from high computational complexity and resource requirements. The quadratic complexity of the self-attention mechanism further exacerbates these challenges when dealing with long sequences and large datasets. Specialized AI hardware accelerators, such as the Habana GAUDI architecture, offer a promising solution to tackle these issues. GAUDI features a Matrix Multiplication Engine (MME) and a cluster of fully programmable Tensor Processing Cores (TPC). This paper explores the untapped potential of using GAUDI processors to accelerate Transformer-based models, addressing key challenges in the process. Firstly, we provide a comprehensive performance comparison between the MME and TPC components, illuminating their relative strengths and weaknesses. Secondly, we explore strategies to optimize MME and TPC utilization, offering practical insights to enhance computational efficiency. Thirdly, we evaluate the performance of Transformers on GAUDI, particularly in handling long sequences and uncovering performance bottlenecks. Lastly, we evaluate the end-to-end performance of two Transformer-based large language models (LLM) on GAUDI. The contributions of this work encompass practical insights for practitioners and researchers alike. We delve into GAUDI's capabilities for Transformers through systematic profiling, analysis, and optimization exploration. Our study bridges a research gap and offers a roadmap for optimizing Transformer-based model training on the GAUDI architecture.
\end{abstract}

\maketitle

\section{Introduction}
\label{sec:introduction}
Transformers have emerged as a powerful and versatile model architecture for various machine learning tasks, particularly in natural language processing (NLP) and visual recognition \cite{vaswani2017attention, devlin2018bert, dosovitskiy2020image}. Despite their impressive ability, Transformers are computationally demanding, which has led to increased concerns regarding their energy efficiency, memory footprint, and deployment cost, especially in real-world applications \cite{wang2022lightseq2,thorpe2023bamboo,9810097}. The quadratic complexity of the self-attention mechanism in Transformers makes it challenging to scale them to long sequences and large-scale datasets, further exacerbating the computational burden.

To address these challenges, researchers have proposed several techniques to improve the efficiency of Transformers. Model compression techniques, such as pruning \cite{han2015deep}, quantization \cite{jacob2018quantization}, and knowledge distillation \cite{hinton2015distilling}, have been explored to reduce the model size and computational requirements while maintaining acceptable performance. Other approaches focus on developing novel attention mechanisms that reduce the computational complexity of the self-attention operation. These include sparse attention \cite{child2019generating}, low-rank attention \cite{wang2020linformer}, and linearized attention \cite{katharopoulos2020transformers}, among others. However, these methods often introduce trade-offs in terms of model accuracy, hardware utilization, and generalization capabilities.

One promising approach to accelerate Transformer computation is the use of specialized hardware accelerators, such as Habana's GAUDI processor \cite{medina2020habana}. The GAUDI architecture is specifically designed for deep learning training workloads and offers a heterogeneous compute architecture comprising a Matrix Multiplication Engine (MME) and a cluster of fully programmable Tensor Processing Cores (TPC). This combination allows GAUDI to efficiently handle various deep learning operations, both matrix-based and non-matrix-based, with high performance and flexibility.

Despite the potential showcased by GAUDI, there exist pivotal gaps in our understanding of its true performance capabilities for Transformer-based models. We outline the multifaceted challenges when accelerating Transformers using GAUDI processors.  
(1) Uncharted performance comparison between MME and TPC. The absence of prior work that comprehensively elucidates the performance comparison between the MME and TPCs within the Habana GAUDI architecture. 
(2) Imbalanced workload for MME and TPC utilization. Optimal workload distribution between the MME and TPC emerges as a decisive factor in maximizing GAUDI's efficacy for Transformers. There is no prior work to address this challenge. Our research delves into this territory, exploring strategies to intricately balance the tasks assigned to MME and TPC. 
(3) Unexplored Transformer performance in long sequences. The third challenge pertains to the performance of Transformers on Habana's GAUDI, particularly in scenarios involving long input sequences. This uncharted territory lacks exploration, hindering our ability to grasp the GAUDI's prowess in handling extended sequences. 
(4) Lack of end-to-end large language model (LLM) performance on GAUDI. The dearth of existing research in a holistic evaluation of end-to-end LLM performance on Habana's GAUDI, coupled with an exploration of potential performance bottlenecks. 

To address those challenges, we benchmark and analyze deeply the performance of Transformers and Transformer-based models on Habana's GAUDI.

The main contributions of this paper are summarized as follows:
\begin{itemize}
  \item We conduct an in-depth performance comparison between the Matrix Multiplication Engine (MME) and Tensor Processing Cores (TPC) within GAUDI. Our analysis offers insights into the relative strengths and weaknesses of these components, empowering practitioners to make informed decisions when tailoring Transformers to the GAUDI platform.
  \item We explore strategies to balance the workload effectively between MME and TPC, we provide practical guidance to achieve enhanced performance and efficiency for Transformers on GAUDI.
  \item We tackle the dearth of research in evaluating the performance of Transformers on GAUDI, especially when dealing with long sequences. Through systematic benchmarking and analysis, we uncover the performance bottlenecks that arise in this scenario, shedding light on the unique challenges posed by long input sequences.
  \item We assess the overall performance of Transformer-based models on Habana's GAUDI and identify performance bottlenecks, we offer a holistic perspective on GAUDI's capabilities for accelerating complex language models.
\end{itemize}

In summary, through this comprehensive study, our work demonstrates the potential of specialized hardware accelerators like GAUDI processors. We contribute a deeper understanding of Habana's GAUDI for Transformers and Transformer-based models. Our findings not only address existing research gaps but also provide practitioners and researchers with valuable insights to optimize the performance of Transformers and Transformer-based models on GAUDI, further unlocking the potential of these models for real-world applications.

\section{Background and Motivation}
\label{sec:background}
In this section, we present background information on the Habana GAUDI processor architecture, the TPC programming model, Transformers, and our motivation.

\subsection{Habana GAUDI Processor Architecture}
\label{subsec:gaudi}
\begin{figure}[h]
    \centering
    \includegraphics[width=0.7\linewidth]{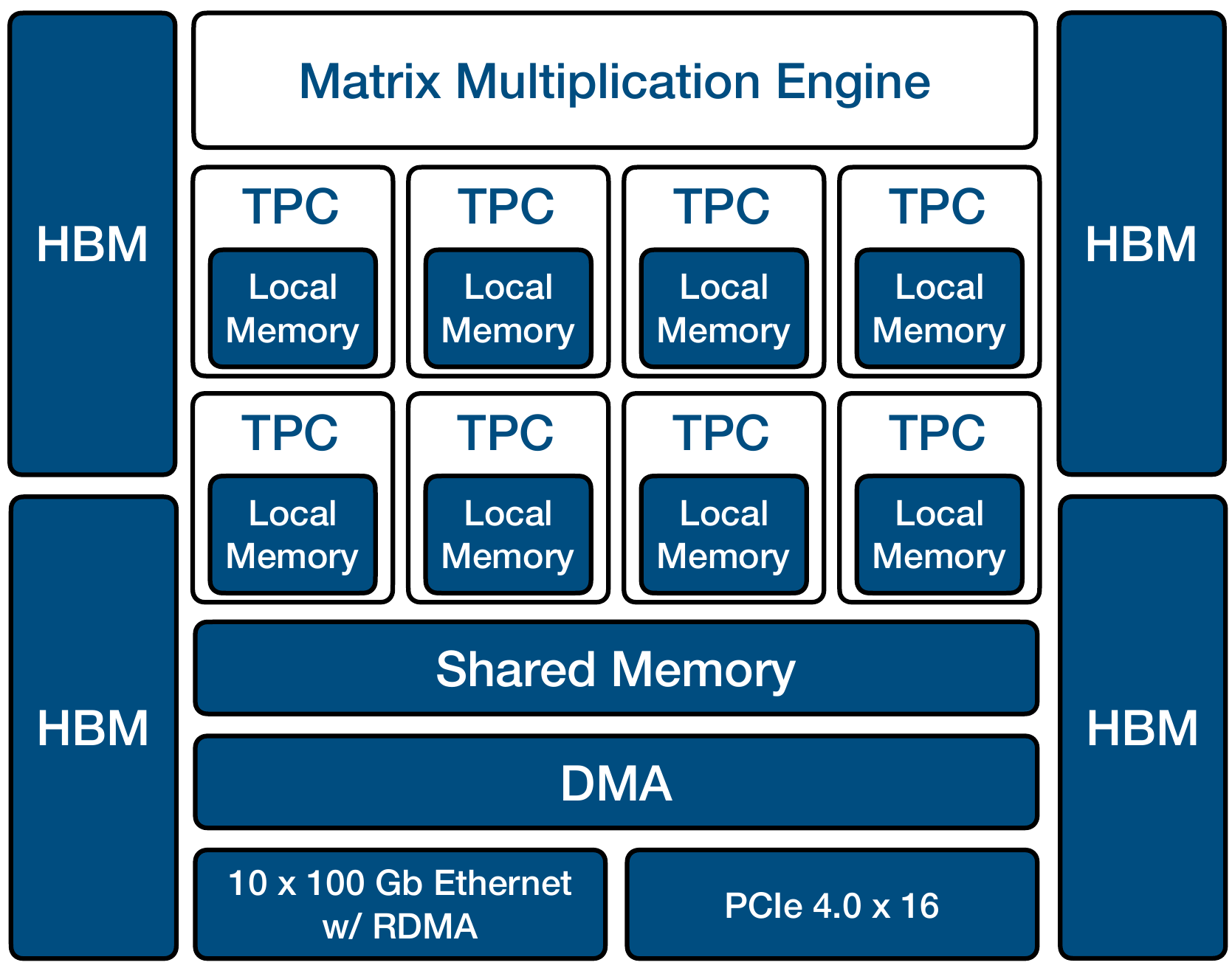}
    \vspace{-2mm}
    \caption{A high-level view of GAUDI architecture, which consists of Matrix Multiplication Engine (MME), Tensor Processing Cores (TPC), Memory Units (Local Memory, Shared Memory, DMA, HBM, RDMA), and Connection Units (Ethernet, PCIe).}
    \label{fig:overview}
   \vspace{-4mm}
\end{figure}

Habana GAUDI processor is a specialized hardware accelerator designed for deep learning training workloads \cite{medina2020habana}. As shown in Figure \ref{fig:overview}, it features a heterogeneous compute architecture with a Matrix Multiplication Engine (MME), eight fully programmable Tensor Processing Cores (TPC), and fast memory and network units. Specifically, GAUDI efficiently handles various deep learning operations by lowering them into matrix multiplication operations (e.g., convolution) and nonlinear operations (e.g., activation) that can be executed on MME and TPC, respectively. The fast memory and network units enhance intra-/inter- processor data transfers, respectively.

MME is specifically tuned for computation tasks in deep neural network (DNN) training such as fully connected layers, convolution layers, and batched-GEMM, providing significant acceleration compared to traditional CPU and GPU solutions~\cite{HabanaWhitePap}. The TPC is a very long instruction word (VLIW) single instruction multiple data (SIMD) processor crafted for deep learning nonlinear operations. It is designed to accelerate non-matrix-based operations that cannot be efficiently handled by the MME. The programming approach of TPC offers users a high degree of flexibility and innovation, supported by features tailored to various workloads. These include acceleration for non-GEMM operations, tensor-based addressing, capabilities to hide latency, random number production, and advanced implementation of special functions.

GAUDI incorporates a DMA engine, streamlining the data exchange between MME and TPC using shared memory. For communications between different processors, GAUDI includes on-chip RoCE v2 engines, facilitating efficient inter-processor dialogue during training sessions. Consequently, GAUDI ensures seamless collaboration between MME and TPC and delivers exceptional scalability in both expanding and multiplying setups.

\begin{figure}[h]
    \centering
    \includegraphics[width=0.7\linewidth]{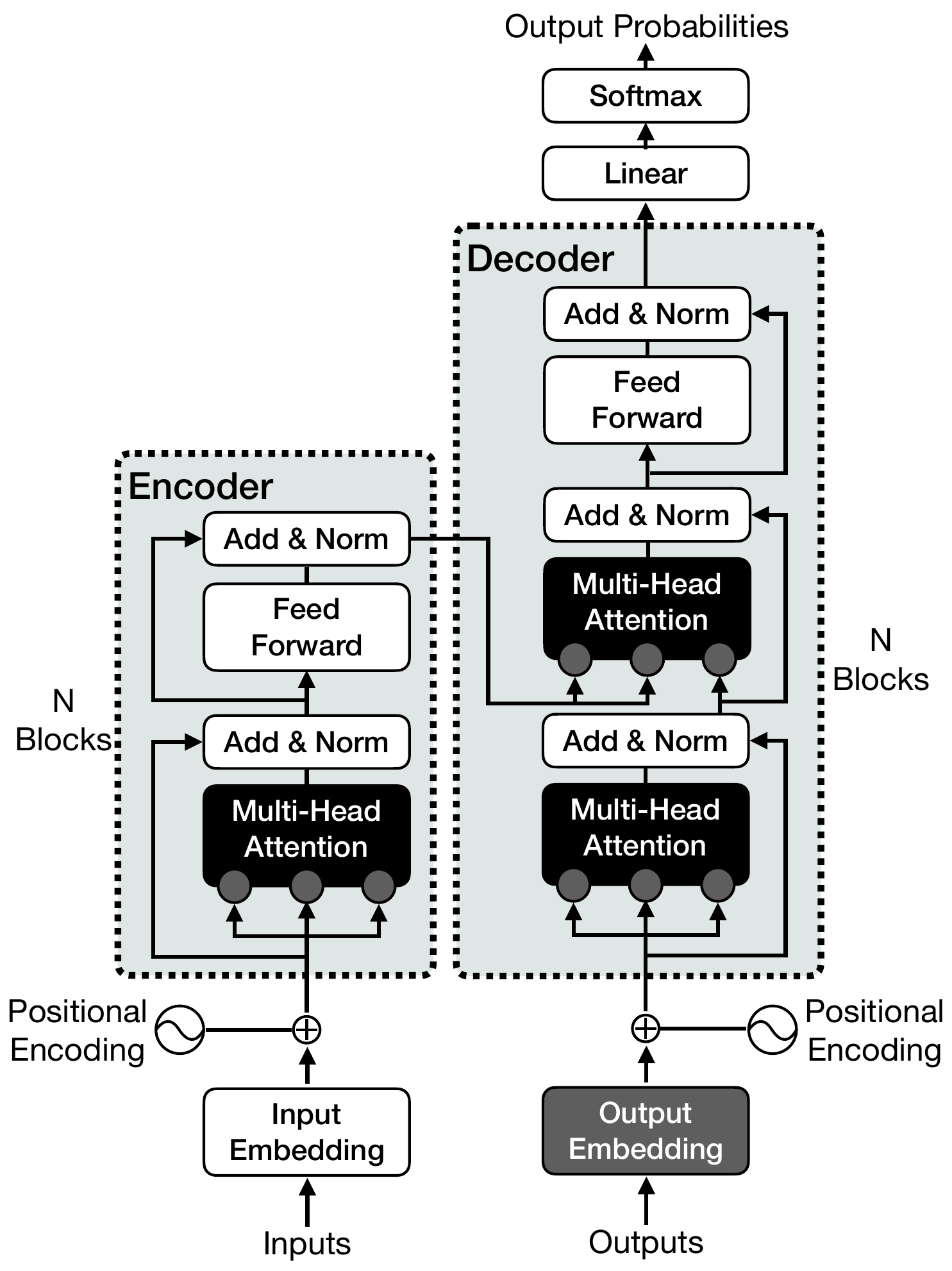}
    \vspace{-2mm}
    \caption{Transformer model architecture overview, which mainly consists of multi-head attention.}
    \label{fig:transformer}
   \vspace{-4mm}
\end{figure}

\subsection{TPC programming model}
\paragraph{\textbf{TPC architecture}}
The TPC boasts a very long instruction word (VLIW) design. Its wide single instruction multiple data (SIMD) vector mechanism can handle 2048-bit SIMD tasks and is compatible with several data types like float, bfloat16, INT16, INT32, and INT8. The instruction set for the TPC processor is segmented into four functional slots:
\begin{itemize}
    \item Load slot - responsible for memory loading, value movements, and value settings.
    \item SPU slot - handles scalar computations.
    \item VPU slot - manages vector computations.
    \item Store slot - oversees memory storage, value movements, and value settings.
\end{itemize}
Four distinct memory domains are embedded within the TPC processor: scalar local memory, vector local memory, global memory, and configuration space. The global memory can be interfaced through specialized access points termed as tensors. On average, every four cycles can accommodate the loading or writing of a 2,048-bit vector to the global memory. It's also worth noting that individual TPC maintain distinct local memory instances, and each TPC can exclusively access its dedicated local cache. The local memory is bifurcated into two storage banks, scalar local memory (1 KB) and vector local memory (80 KB). There's an unrestricted bandwidth when reading from or writing to the local memory in each cycle.

\paragraph{\textbf{TPC programming}}
The TPC stands as a fully programmable VLIW SIMD processor, programable via TPC-C, a C language derivative. TPC-C incorporates vector data types for seamless use of processor-specific SIMD capabilities. A TPC program is composed of host glue code and a TPC kernel. Host glue code, executed on the host machine, controls program execution. TPC kernels, executed on TPC processors, handle computation. Users leverage the SynapseAI TPC SDK, featuring an LLVM-based TPC-C compiler, simulator, and debugger, for TPC kernel development. 
The TPC processor on the GAUDI ASIC accepts tensor inputs/outputs with dimensions ranging from 1 to 5. Index spacing, similar to threads in CUDA programming, efficiently divides workloads among TPC processors. Each index space member corresponds to an independent unit of work executed on a single TPC. Users utilize Habana's intrinsics, encompassing arithmetic, bitwise, and load operations, to create TPC kernels, while ensuring effective workload distribution.

\begin{figure}[t]
    \centering
    \includegraphics[width=0.8\linewidth]{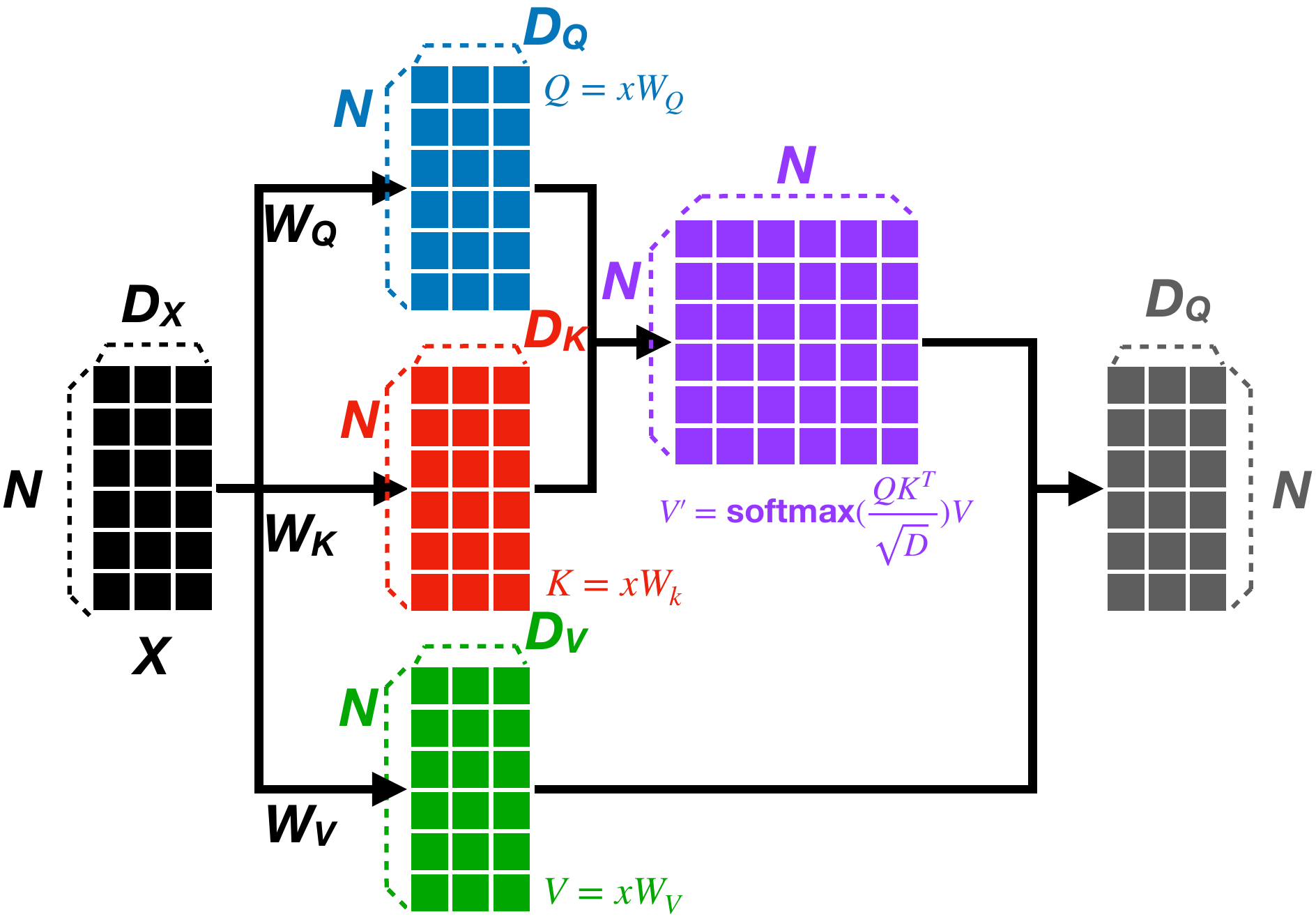}
    \vspace{-2mm}
    \caption{Matrix Computation workflow of each self-attention. $Q$, $K$ and $V$ are query, key, value matrices of dimension size $N$ by $D_Q$,$D_K$, $D_V$, respectively.}
    \label{fig:matrix_computation}
   \vspace{-4mm}
\end{figure}

\subsection{Transformers}
The Transformer architecture was first introduced by Vaswani et al. \cite{vaswani2017attention} as a novel approach to sequence-to-sequence learning tasks, particularly in natural language processing. Transformers have since become a popular choice for various machine-learning applications, including language modeling, machine translation, and computer vision. 
The key innovation of the Transformer architecture is the self-attention mechanism, which allows the model to weigh different parts of the input sequence differently when making predictions. This mechanism enables Transformers to capture long-range dependencies and contextual information more effectively compared to traditional recurrent neural networks (RNNs) and convolutional neural networks (CNNs). 
Figure \ref{fig:transformer} presents the architecture of a Transformer, which typically consists of encoder blocks, decoder blocks, and other operations such as position embedding and layer normalization. Specifically, each encoder/decoder block consists of multi-head self-attention mechanisms followed by a position-wise feed-forward network. Many widely-received DNN models are based on Transformers. For example, the Bidirectional Encoder Representations from Transformers (BERT) \cite{devlin2018bert} and the Generative Pre-trained Transformer (GPT) \cite{radford2018improving}. BERT is primarily an encoder from the Transformer architecture. GPT is both an encoder and a decoder, but during training, only the decoder portion is utilized. BERT is bidirectional, trying to understand the context on both sides of a word. GPT is unidirectional, predicting words based on the preceding context.

\begin{table}[h]
    \caption{Operation-Hardware Mapping via SynapseAI}
    \centering
    \resizebox{0.8\linewidth}{!}{
        \begin{tabular}{rrrrr}
\toprule
  \textbf{\sffamily Operation}
& \textbf{\sffamily Explanation}
& \textbf{\sffamily Mapping} \\
\midrule
torch.mul         &element wise mul   &TPC \\
torch.matmul      &matrix product     &MME \\
torch.square      &tensor square      &TPC \\
**                &tensor square      &TPC \\
tensor +- tensor  &tensor +- tensor   &TPC \\ 
scalar * tensor   &scalar * tensor    &TPC \\ 
scalar +- tensor  &scalar +- tensor   &TPC \\ 
torch.sqrt        &square root        &TPC \\ 
torch.log         &natural logarithm  &TPC \\ 
\bottomrule
\end{tabular}
    }
    \label{tab:operations}
    \vspace{-4mm}
\end{table}

\subsection{Motivation}
The impressive ability of Transformer-based models comes from complex computational operations and the huge number of parameters (340 million in BERT, 1.5 billion in GPT-3) \cite{devlin2018bert,brown2020language}, which results in intensive computations during training. Consequently, training Transformer-based models is both time-consuming and resource-intensive. Although today's AI accelerators, such as Habana GAUDI outperform GPUs in some training tasks \cite{HabanaWhitePap}, the architecture-specific optimizations on these accelerators are not well studied. 
For example, Figure \ref{fig:matrix_computation} shows the workflow of matrix computations in self-attention. Specifically, The input sequence $x \in \R ^{N \times D_x}$ is projected by three weight matrices $W_Q, W_K, W_V$ to corresponding representations $Q$, $K$ and $V$. Following common terminology, the $Q$, $K$, and $V$ are referred to as the "queries", "keys", and "values" respectively. Then softmax is used to normalize attention matrix $QK^T$ into a probability distribution. The softmax's computation can only be executed on TPC, which degrades the overall training performance of Habana GAUDI (to be detailed in \S \ref{sec:experimental}). Thus, we perform comprehensive profiling on Habana GAUDI with insights that derive our optimizations in improving the training performance. 

\section{Experimental Results}
\label{sec:experimental}
In this section, we present our experimental setup, profiling results, and discussion. 

\subsection{Experimental Setup}
\vspace{-1mm}
\paragraph{\textbf{Platforms}}
We perform our experiments on one Habana Labs System 1 (HLS-1) \cite{medina2020habana} AI training system. The HLS-1 incorporates eight GAUDI processors and two Gen 4.0 PCIe switches. External Host CPU is used to manage HLS-1 via PCIe switches. Each GAUDI processor is equipped with 32 GB on-chip memory. All experiments are on a single GAUDI processor.

\vspace{-1mm}
\paragraph{\textbf{Implementation details}}
Habana’s SynapseAI \cite{medina2020habana} software suite enables efficient mapping of neural network topologies onto GAUDI hardware. All experiments are performed on PyTorch-based SynapseAI. The PyTorch version is 1.13.1. 

\subsection{Basic Profiling}
\label{subsec:basic}
\vspace{-1mm}
\paragraph{\textbf{Operation mapping}}
PyTorch provides a variety of operations. GAUDI's compute architecture is heterogeneous and includes two independent compute engines – an MME and a fully programmable TPC cluster. So it is necessary for us to know which compute engine each operation is finally mapped to. We perform detailed profiling to obtain the operation-compute engine mapping, as shown in Table \ref{tab:operations}. From this table, we draw the following conclusions: only matrix multiplication operations are mapped to MME, and all other operations are mapped to TPC. Even linear operations on tensors like tensor multiplied by scalar are mapped to TPC. 

\begin{figure*}[ht!]
    \centering
    \includegraphics[width=1\linewidth]{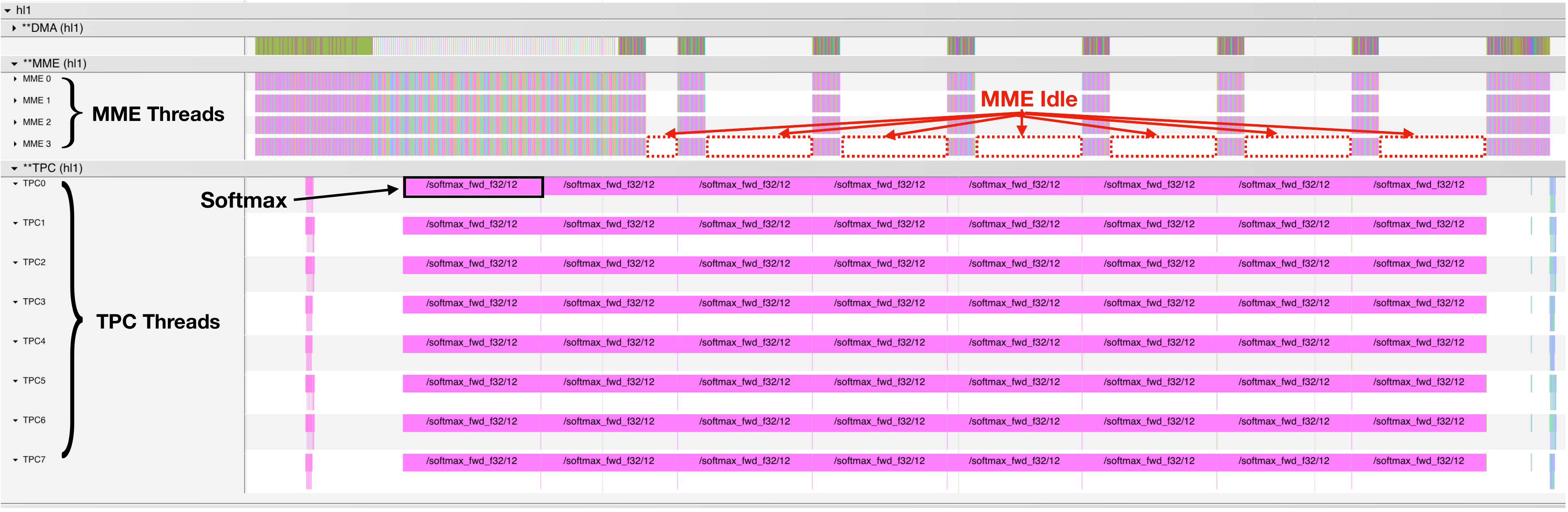}
    \vspace{-2mm}
    \caption{Profiler Trace of the transformer with softmax attention. DMA is direct memory access engine that manages data transfer/copy between MME and TPC. We observe that executing softmax operations on TPC results in MME idle time (i.e., gaps between MME operations).}
    \label{fig:softmax_attn}
   \vspace{-4mm}
\end{figure*}

\vspace{-1mm}
\paragraph{\textbf{Performance comparison between MME and TPC}}
A detailed performance comparison between MME and TPC is very necessary because it helps us analyze the performance bottleneck of the GAUDI. Different operations in the neural network will either be mapped to MME or TPC, and the slowest operation on two compute engines will become a performance bottleneck.

To profile computation performance, we enable MME and TPC to perform batch matrix multiplication operations on various dense matrices of different sizes and measure the run time and tera floating point operations per second (TFLOPS). We directly choose \texttt{torch.bmm} on MME to perform a batch matrix-matrix product, where the batch size is set to 64. We implement TPC batch matrix-matrix product kernels using example code from Habana\_Custom\_Kernel repository \cite{CustomKernel}. SynapseAI profiler is used as suggested by Habana to generate hardware trace events and accurately measure the execution time of each operation. Table \ref{tab:matrix_multi} shows the execution time between MME and TPC for matrix multiplications of different sizes. We can conclude that the computational performance of TPC is up to 7$\times$ lower than that of MME. In the case of such an obvious performance gap, the most suitable application scenario for GAUDI is that the current operation has a large amount of calculation and can be successfully mapped to MME. The next operation has a small amount of calculation and can be mapped to TPC, in such a situation TPC will not form a computing performance bottleneck. But if the next operation has a similar amount of calculation, then MME has to become idle and wait for the calculation of TPC to complete.

\begin{table}[t]
    \caption{Comparison of execution time between MME and TPC for matrix multiplication of different sizes. T\_MME, F\_MME, T\_TPC, F\_TPC are short for run time of MME, TFLOPS of MME, run time of TPC, TFLOPS of TPC, respectively. $\text{Speedup}=\frac{\text{T\_TPC}}{\text{T\_MME}}$. Time unit is millisecond (ms).}
    \centering
    \resizebox{0.8\linewidth}{!}{
        \begin{tabular}{rrrrrrr}
\toprule
  \textbf{\sffamily Size}
& \textbf{\sffamily T\_MME}
& \textbf{\sffamily F\_MME} 
& \textbf{\sffamily T\_TPC}
& \textbf{\sffamily F\_TPC} 
& \textbf{\sffamily Speedup} \\
\midrule
128  &7.31   &2.35  &9.21    &1.86 &1.3 \\
256  &11.78  &11.67 &67.04   &2.05 &5.7 \\
512  &76.51  &14.37 &516.60  &2.13 &6.7 \\
1024 &151.03 &14.56 &1006.30 &2.18 &6.7 \\
2048 &338.27 &14.59 &2247.80 &2.19 &6.6 \\
\bottomrule
\end{tabular}

    }
    \label{tab:matrix_multi}
    \vspace{-4mm}
\end{table}

\subsection{Transformer Layer Profiling}
\label{subsec:layer_prof}

\vspace{-1mm}
\paragraph{\textbf{Softmax attention}}
Self-attention computes, for every position, a weighted average of the feature representations of all other positions with a weight proportional to a similarity score between the representations. Transformers usually follow original design \cite{vaswani2017attention} by Ashish Vaswani to adopt softmax attention. Softmax attention is a specific form of self-attention where the similarity score is the exponential of the dot product between a query and a key. Similarity function is $sim(q, k)=exp(\frac{q^T K}{\sqrt{D}})$. The $Q$, $K$, and $V$ are referred to as the "queries", "keys", and "values" respectively.

Long sequence training in Transformer-based natural language processing (NLP) models, such as BERT and GPT, offers several significant benefits: (1). Capturing long-range dependencies: Long sequence training allows Transformer models to capture these complex dependencies, enabling a better understanding of the context and improving the quality of language representations. (2). Improved contextual understanding: Longer sequences provide more context to the model, allowing it to comprehend the nuances and subtleties in language. (3). enhanced text generation: Longer context windows help the model maintain better coherence and consistency in longer text generation tasks. (4). Better handling of large documents: In real-world applications, NLP models often encounter long documents or lengthy pieces of text. 
Because of the advantages of long sequence training, in experiments, we set the input sequence length, batch size, the number of heads, and the hidden size per head as 2048, 128, 6, and 64 respectively.

Figure \ref{fig:softmax_attn} shows a profiling result of a single Transformer layer. From this result, we have two observations. (1). There are many blank areas in the MME operating area. These blank areas indicate that MME is idle is waiting for tasks. (2). In the running region of TPC, it is very clearly shown that the running time of softmax exceeds 80\% of the total running time.

The reasons for this phenomenon are: (1). The TPC is less computationally powerful than the MME as discussed in Section \ref{subsec:basic}. But The computational complexity of softmax operation in a Transformer is $\mathcal{O}(N^2)$. As a result, it becomes a performance bottleneck when softmax operation is mapped into TPC. (2). Softmax requires reduction operations, which are not well-suited for single instruction, multiple data (SIMD) architectures like TPC. Long sequences further exacerbate this problem especially when the sequence length exceeds 1024. Overall, the limited computational capability of TPC combined with the complexities of softmax operations on this architecture hinders GAUDI's overall performance and efficiency.

\vspace{-1mm}
\paragraph{\textbf{Linearized attention}}
Linearized attention, also known as "linear attention", is an alternative approach to the traditional softmax-based attention mechanism used in Transformers. It aims to reduce the computational complexity associated with the softmax operation while maintaining the core principles of self-attention. Linear attention is particularly useful when dealing with very long sequences, where standard softmax-based self-attention becomes impractical due to its quadratic complexity.

The softmax-based self attention is $\text{softmax}(\frac{QK^T}{\sqrt{D}})V$, where $Q, K$ and $ V \in \R ^{N \times D}$. The computational complexity of self-attention is quadratic to sequence length $N$. Assume $\phi$ is a feature map that is applied in a row-wise manner, linear attention is $(\phi(Q)\phi(K)^T)V = \phi(Q)(\phi(K)^T V)$ after applying the associative property of matrix multiplication. linear attention leads to a computational complexity of $\mathcal{O}(N)$.

There are two reasons why we want to use linear attention on Habana: (1). The calculation of the softmax operation itself is relatively complicated, and it involves exponential operations and reduction operations. (2). The essence of linear attention is that matrix multiplication can ensure that almost all self-attention calculations are mapped to MME with stronger computation performance.

\begin{lstlisting}[language=Python,style=mystyle,linewidth=\linewidth,caption=Pseudocode for FAVOR Algorithm.]
def FAVOR(q,k,v):
    #Project key and queries to feature map space
    q_scaled = self.pre_scale(q)
    q_scaled = q_scaled @ self.features
    q_prime = torch.exp(q_scaled + self.offset)
    k_scaled = self.pre_scale(k)
    k_scaled = k_scaled @ self.features
    k_prime = torch.exp(k_scaled + self.offset)
    att_norm = q_prime @ (
        k_prime.transpose(-2,-1) @ torch.ones_like(v)
        )
    att_raw = q_prime @ (k_prime.transpose(-2,-1) @ v)
    x = att_raw / att_norm
    return x
\end{lstlisting}
\vspace{-2mm}

We adopt feature maps from Linear Transformers \cite{katharopoulos2020transformers} and Performer \cite{choromanski2020rethinking} to construct linear attention on Habana. Linear Transformer proposes to directly set the feature map as $\phi(x) = elu(x) + 1$. 
The Performer uses a novel Fast Attention Via a positive Orthogonal Random features approach (FAVOR). Its feature map is $\phi(x) = \frac{h(x)}{\sqrt{m}}(f_1(\omega_1^T x), \cdots, f_1(\omega_m^T x), \cdots, f_l(\omega_1^T x), \cdots, f_l(\omega_m^T x))$, where $f_1, \cdots, f_l: \R \rightarrow \R $. $\omega_1, \cdots, \omega_m$ are drawn from some distribution. 

\begin{figure}[h]
    \centering
    \includegraphics[width=\linewidth]{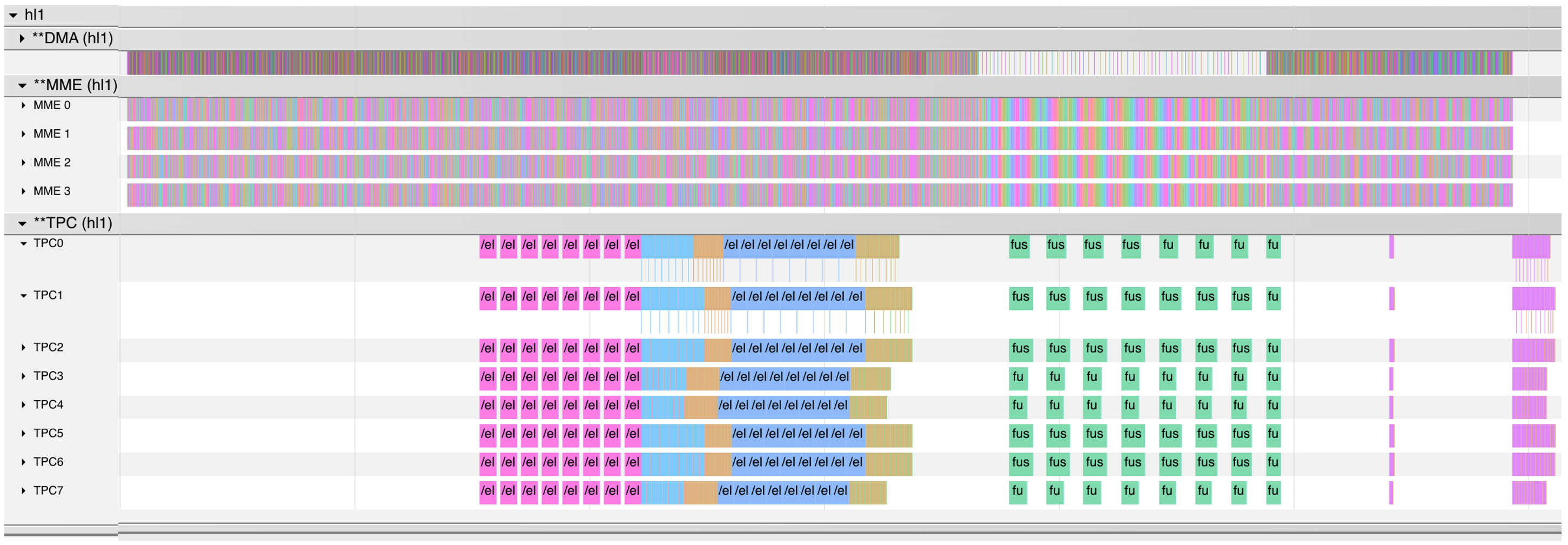}
    \vspace{-4mm}
    \caption{Profiling of linear Transformers. Colored blocks are computation periods and gaps between colored blocks are idle periods.}
    \label{fig:linear_attn}
    \vspace{-2mm}
\end{figure}

\begin{figure}[h]
    \centering
    \includegraphics[width=\linewidth]{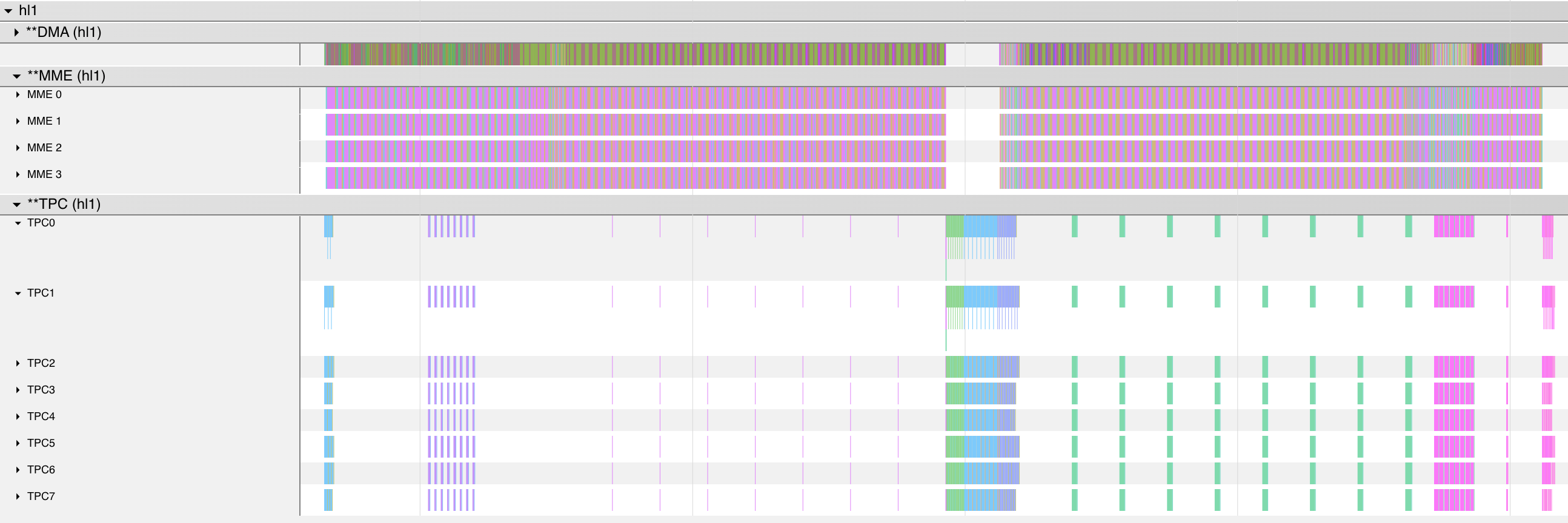}
    \vspace{-2mm}
    \caption{Profiling of Performer. Colored blocks are computation periods, and gaps between colored blocks are idle periods.}
    \label{fig:favor_attn}
   \vspace{-4mm}
\end{figure}

Figure \ref{fig:linear_attn} \ref{fig:favor_attn} depicts profiling results of linear Transformers and Performers. The total run time of linear Transformers and Performer is 30 ms and 80 ms, respectively. Compared to original softmax-based attention, linear Transformers and Performer achieve 6 $\times$, 2 $\times$ speedup. Besides there are not many blank areas in the MME operating area, which indicates full utilization of MME. Therefore, we can conclude that linearized attention is a good alternative to softmax attention from the perspective of performance.

However, there is a blank area in the MME operating area when using Performer. The blank area is because the TPC is busy with exponential operations. As shown in the algorithm of FAVOR, we can find that the calculation of "q\_prime" and "k\_prime" is independent. But Graph Compiler does not detect this independence, so it does not schedule MME and TPC tasks well so that they can overlap.

\paragraph{\textbf{Activation functions}}
Linear Transformer \cite{katharopoulos2020transformers} does not consider the impact of different activation functions on TPC performance, it directly sets the activation function to exponential linear unit (ELU). And there is no previous work discussing the performance of different activation functions on TPC. Thus we conduct a rigorous evaluation to assess the impact of various activation functions on the overall efficiency and computational capabilities of the TPC. The experiments incorporate popular activation functions explored in NLP tasks, including rectified linear unit (ReLU), LeakyReLU, Gaussian Error Linear Units (GELU), and gated linear unit function (GLU). 

\begin{figure*}[h]
    \centering
    \includegraphics[width=\linewidth]{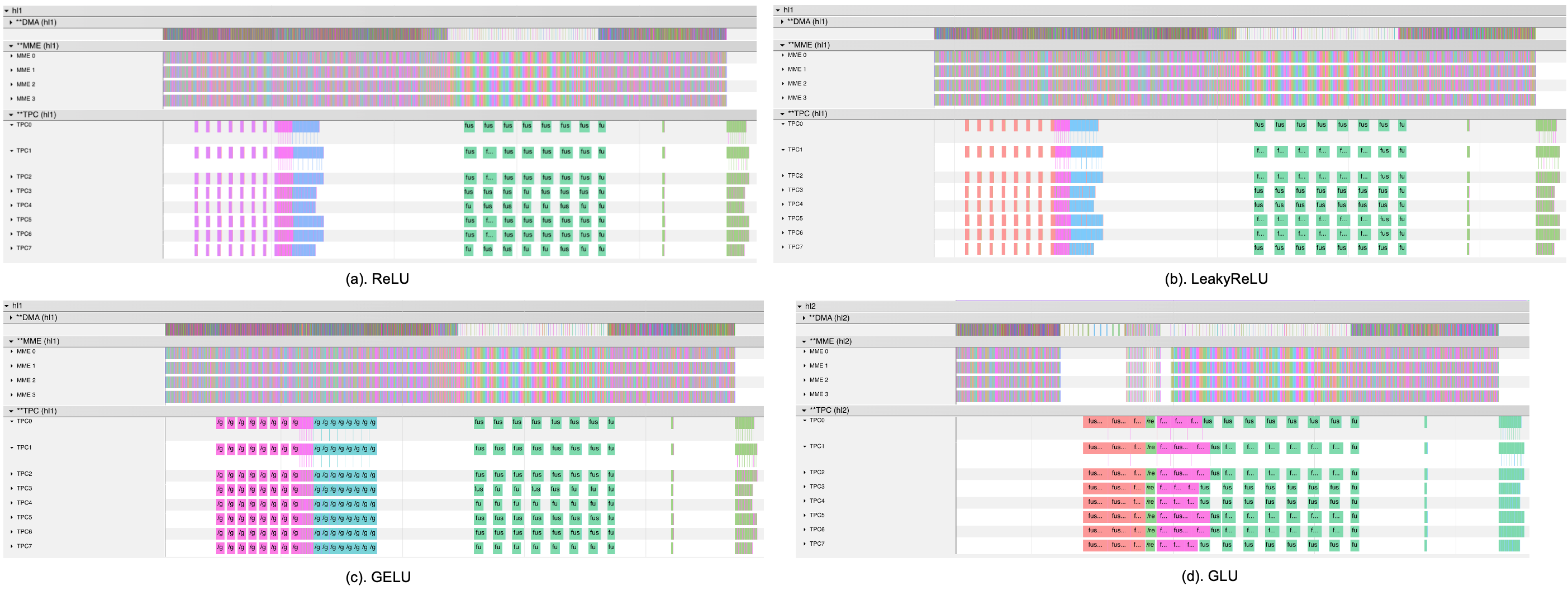}
    \vspace{-2mm}
    \caption{Activation functions in NLP. }
    \label{fig:act_funcs}
   \vspace{-2mm}
\end{figure*}

In experiments, we set the input sequence length, batch size, the number of heads, and the hidden size per head to 2048, 128, 6, and 64 respectively. 
Figure \ref{fig:act_funcs} depicts hardware traces of different activation functions. From the profiling results, we have two observations 1. The total run time of a Transformer with ReLU, LeakyReLU, GELU, and GLU is 30.1 ms, 30.2 ms, 29.7 ms, and 32.6 ms, respectively. Transformers with ReLU, LeakyReLU, and GELU have similar performance and The execution of MME and TPC has a good overlap. 2. Transformer with GLU has the worst performance. And its execution causes a blank area in MME. We think the reasons for such phenomena are (1). those activation functions are applied to element-wise tensor, which is extremely suitable for SIMD architecture like TPC. (2). SynapseAI does not have good support for GLU, which cause extra compilation during the execution. 


\subsection{End-To-End Language Models Profiling.}
In order to analyze the end-to-end performance of a full language model on GAUDI, we choose profile execution of BERT and GPT run on GAUDI. For GPT model, we utilize the GPT2LMHeadModel module from Hugging Face \cite{wolf2020transformers}. GPT2LMHeadModel is the GPT2 Model Transformer with a language modeling head on top. For the BERT model, we use the BertForMaskedLM module from Hugging Face. BertForMaskedLM is the BERT model with a language modeling head on top. The input dataset is book corpus \cite{zhu2015aligning}. Due to limited GAUDI memory, we set the input sequence length, batch size, the number of layers, the number of heads, and the hidden size per head as 2048, 8, 2, 8, and 64 respectively.

\begin{figure*}[ht]
    \centering
    \includegraphics[width=0.8\linewidth]{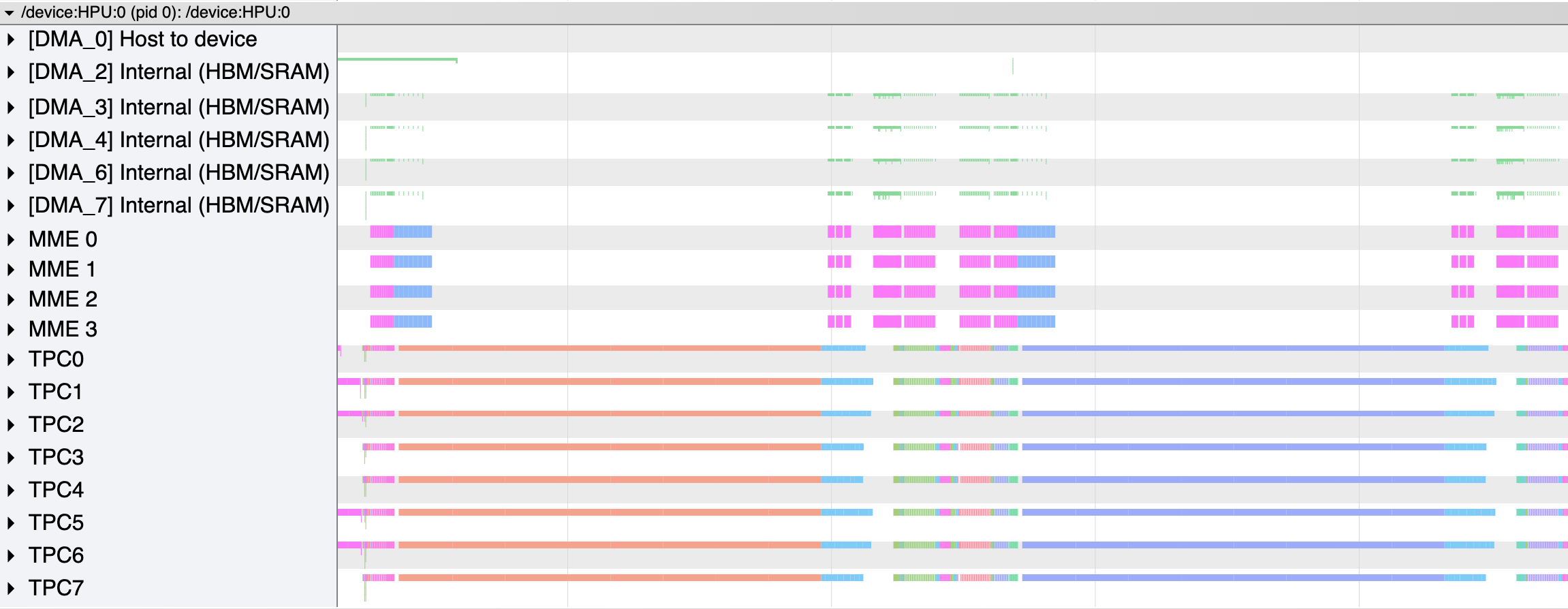}
    \vspace{-2mm}
    \caption{Hardware trace of GPT model.}
    \label{fig:gpt_2l_2048pos}
\end{figure*}

\begin{figure*}[ht]
    \centering
    \includegraphics[width=0.8\linewidth]{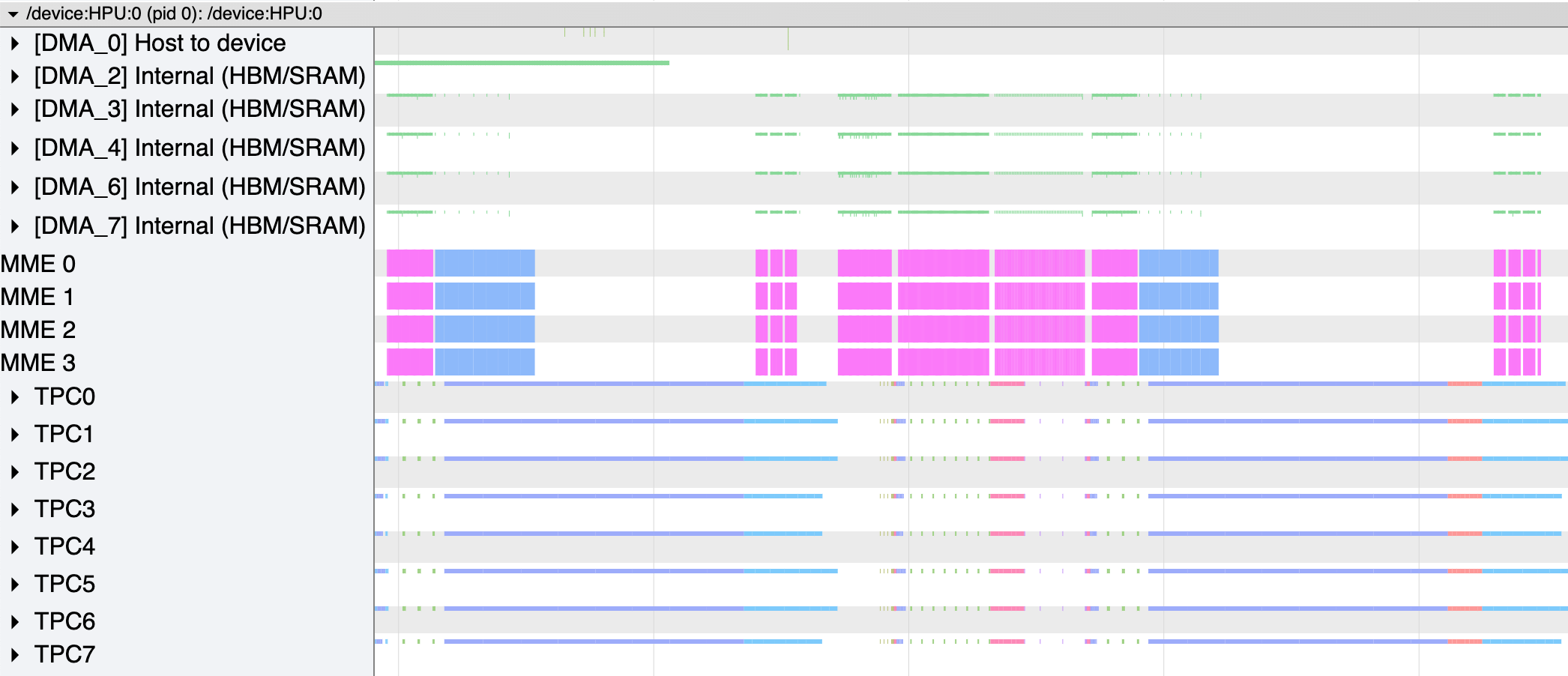}
    \vspace{-2mm}
    \caption{Hardware trace of BERT model.}
    \label{fig:bert_2l_2048pos}
\end{figure*}

Figure \ref{fig:gpt_2l_2048pos}, \ref{fig:bert_2l_2048pos} show hardware traces of GPT and BERT models. From traces, we have similar observations as single Transformer layer profiling. There are many blank areas in the MME operating area, which indicates MME is idle. However, TPC is obviously busy. Potential performance issues of Transformer-based language models on GAUDI are (1). workload between MME and TPC is unbalanced. (2). There is no good overlap between MME and TPC. As a result, either MME or TPC is ideal, which causes a waste of computing resources.

\section{Insights and Takeaways}
\label{sec:insights}
(1) We need to try to provide all source code so GraphCompiler can analyze the source code thoroughly and generate good mapping and schedule of MME and TPC. 
(2) The code should use very basic operations provided by Torch and avoid high-level abstracts like torch.einsum() for good mapping and schedule of MME and TPC by GraphCompiler. 
(3) When designing a neural network model, the user should consider that most calculations in the model can be transformed into matrix multiplication. In this way, the model can fully utilize MME' powerful computation capability.

\section{Conclusion and Future Work}
\label{sec:conclusion}
In this work, we embarked on a comprehensive exploration of the performance capabilities of Habana's GAUDI processor when accelerating Transformers and Transformer-based models. 
Our findings not only address existing research gaps but also provide practitioners and researchers with valuable insights to optimize the performance of Transformers and Transformer-based models on GAUDI, further unlocking the potential of these models for real-world applications.
In the future, we plan to investigate novel attention mechanisms tailored to GAUDI's architecture could also optimize performance for long sequences. 

\begin{acks}
\small The material was supported by the U.S. DOE Office of Science (SC), Office of Advanced Scientific Computing Research (ASCR), under contracts DE-AC02-06CH11357. This work was also supported by NSF awards 2303820, 2303064, 2247080, 2311876, and 2312673. We gratefully acknowledge the computing resources provided and operated by the Joint Laboratory for System Evaluation (JLSE) at Argonne National Laboratory.
\end{acks}

\clearpage
\bibliographystyle{ACM-Reference-Format}
\bibliography{refs}

\end{document}